\documentclass[conference]{IEEEtran}
\IEEEoverridecommandlockouts
\usepackage{cite}
\usepackage{amsmath,amssymb,amsfonts}
\usepackage{graphicx}
\usepackage{textcomp}
\usepackage{xcolor}
\def\BibTeX{{\rm B\kern-.05em{\sc i\kern-.025em b}\kern-.08em
    T\kern-.1667em\lower.7ex\hbox{E}\kern-.125emX}}

\usepackage{balance}
\usepackage{nicematrix}
\usepackage{multirow}
\usepackage{booktabs}
\usepackage{algorithm}
\usepackage{algorithmic}
\usepackage{soul}

\usepackage{amsmath}
\usepackage{tcolorbox}
\usepackage{ulem}
\usepackage{color, colortbl}
\definecolor{LightGray}{rgb}{0.78, 0.79, 0.85}
\definecolor{LightCyan}{rgb}{0.,0.5,0.71}

\usepackage[numbers]{natbib}

\begin{document}

\title{Deploying Deep Reinforcement Learning Systems: A Taxonomy of Challenges}

\author{
\IEEEauthorblockN{Ahmed Haj Yahmed}
\textit{ahmed.haj-yahmed@polymtl.ca}
\IEEEauthorblockA{
\textit{Polytechnique Montréal, Canada}}
\and
\IEEEauthorblockN{Altaf Allah Abbassi}
\textit{altaf-allah-2.abbassi@polymtl.ca}
\IEEEauthorblockA{
\textit{Polytechnique Montréal, Canada}}
\and
\IEEEauthorblockN{Amin Nikanjam}
\textit{amin.nikanjam@polymtl.ca}
\IEEEauthorblockA{
\textit{Polytechnique Montréal, Canada}}
\and
\IEEEauthorblockN{Heng Li}
\textit{heng.li@polymtl.ca}
\IEEEauthorblockA{
\textit{Polytechnique Montr\'{e}al, Canada}}
\and
\IEEEauthorblockN{Foutse Khomh}
\textit{foutse.khomh@polymtl.ca}
\IEEEauthorblockA{
\textit{Polytechnique Montr\'{e}al, Canada}}
}

\maketitle

\begin{abstract}
Deep reinforcement learning (DRL), leveraging Deep Learning (DL) in reinforcement learning, has shown significant potential in achieving human-level autonomy in a wide range of domains, including robotics, computer vision, and computer games. This potential justifies the enthusiasm and growing interest in DRL in both academia and industry. However, the community currently focuses mostly on the development phase of DRL systems, with little attention devoted to DRL deployment. In this paper, we propose an empirical study on Stack Overflow (SO), the most popular Q\&A forum for developers, to uncover and understand the challenges practitioners faced when deploying DRL systems. Specifically, we categorized relevant SO posts by deployment platforms: server/cloud, mobile/embedded system, browser, and game engine. After filtering and manual analysis, we examined $357$ SO posts about DRL deployment, investigated the current state, and identified the challenges related to deploying DRL systems. Then, we investigate the prevalence and difficulty of these challenges. Results show that the general interest in DRL deployment is growing, confirming the study’s relevance and importance. Results also show that DRL deployment is more difficult than other DRL issues. Additionally, we built a taxonomy of $31$ unique challenges in deploying DRL to different platforms. On all platforms, RL environment-related challenges are the most popular, and communication-related challenges are the most difficult among practitioners. We hope our study inspires future research and helps the community overcome the most common and difficult challenges practitioners face when deploying DRL systems.

\end{abstract}


\section{Introduction}
\label{sec:introduction}
Reinforcement Learning (RL) is a subfield of Machine Learning (ML) concerned with autonomous learning and decision-making based on interacting with an environment \cite{sutton2018reinforcement}. RL follows a trial-and-error paradigm where an agent interacts with its environment and learns to adapt its behavior to achieve a goal by observing the outcomes (i.e., rewards) of its actions \cite{sutton2018reinforcement, li2017deep, arulkumaran2017deep}. RL was at first unpopular since early approaches were constrained to low-dimensional tasks and lacked scalability \cite{sutton2018reinforcement, strehl2006pac}. Deep Reinforcement Learning (DRL), leveraging Deep Learning (DL) in RL, sparked a rebirth in RL and revived interest in this field. It was a step toward developing autonomous systems with a deeper awareness of the surrounding world. DL is currently allowing RL to achieve human-level autonomy in previously intractable fields, such as robotics \cite{8675643}, computer games \cite{DBLP}, and computer vision \cite{panzer2022deep}. 

   This potential explains the enthusiasm and rising interest in DRL in academics and industry. Frameworks and libraries like Stable Baselines \cite{stable_baseline}, Keras-RL \cite{plappert2016kerasrl}, and TensorForce \cite{tensorforce} are continually being released to relieve the cost of building DRL solutions from scratch. Academia and industry are also working together to assist researchers and practitioners handle DRL's new challenges. For example, Nikanjam et al. \cite{nikanjam2022faults} studied real faults that occurred while developing DRL programs and produced a taxonomy of these faults. From another perspective, the growing demand for DRL-based systems has raised new deployment concerns. For instance, these systems' high computational and energy costs prevent their direct deployment on platforms with low processing power (e.g., drone navigation) \cite{shiri2022e2hrl, jagannath2022mr}. Even worse, these additional deployment concerns are generally non-trivial and more difficult compared to vanilla DL deployment. Quantization \cite{cai2017deep, guo2018survey}, for instance, is more challenging in DRL and may hinder the policy's long-term decision-making since the agent's current action strongly affects its future states and actions~\cite{krishnan2019quarl}. However, current research focuses mostly on the development phase of DRL-based systems, with little attention paid to DRL deployment. 

In this paper, we undertake the first attempt at identifying and understanding the challenges practitioners faced when deploying DRL-based software systems. We formulate our Research Questions (RQs) as follows:
 
\begin{itemize}
    \item[-] \textbf{RQ1: What is the current level of interest in deploying DRL-based systems?} 
    \item[-] \textbf{RQ2: What are the challenges in deploying DRL-based systems?} 
    \item[-] \textbf{RQ3: Which DRL deployment challenges are the most popular and difficult to answer on Stack Overflow?} 
\end{itemize}

We conduct an empirical study on Stack Overflow (SO) leveraging a variety of qualitative and quantitative techniques. We investigate SO as it is the most popular Q\&A platform for developers to report their challenges and issues, propose solutions, and spark discussions on various technical topics including DRL \cite{abdalkareem2017developers}. Following similar studies on DL \cite{study, guo2019empirical}, we categorize relevant SO posts by deployment platforms: server/cloud, mobile/embedded system, browser, and game engine. After filtering and manual analysis to remove false positives, we examined 357 SO posts about DRL deployment. Quantitative analysis shows that the general interest in DRL deployment is growing, confirming the study’s relevance and importance. 
We manually review SO posts and build a comprehensive taxonomy of 31 unique challenges for deploying DRL to the selected platforms. 
These 31 deployment challenges can be grouped into 11 main categories. Across all platforms, RL environment-related challenges are the most popular whereas communication-related challenges are the most difficult. Also, when considering average scores and median response time as proxies for popularity and difficulty, we found that the difficult/popular challenges are significantly popular/difficult among practitioners. We found that despite increased interest from the DRL community, DRL deployment needs more work to achieve the same maturity level as traditional software systems deployment. Academia should propose more automated strategies for diagnosing and monitoring deployment issues and misconfigurations to help developers, and framework providers should enhance their tools and documentation.

Our study is important for software maintenance and evolution, like similar studies \cite{chen2021empirical, study, openja2022empirical}, since deployment challenges directly impact the maintenance and evolution of DRL systems in production, for example how a DRL system is deployed can affect its maintenance effort when new changes are needed.

We have prepared a replication package including materials used in this study, that can be used for other studies on DRL deployment \cite{replication}.

The remainder of this paper is as follows: Section \ref{sec:background} outlines DRL system development and deployment. Section \ref{sec:methodology} covers our methodology. Sections \ref{sec:RQ1} to \ref{sec:RQ3} report our empirical findings. Section \ref{sec:Implications} discusses the implications of our study. Section \ref{sec:thre-valid} discusses validity threats, Section \ref{sec:related-work} explores related work, and Section \ref{sec:conclusion} concludes the paper.

\section{Background}
\label{sec:background}
This section briefly discusses DRL-based system development and deployment. Interested readers might turn to~\cite{kurrek2020reinforcement} for a more extensive discussion about a concrete design lifecycle of a DRL-based robot.

\textbf{DRL Development Lifecycle:} The development of DRL-based systems involves four main steps: design, control, training, and verification \cite{kurrek2020reinforcement}. First, developers select and combine the components of the system to construct its final structure (i.e., design step). Then, they start generating modules to control the internal and external behavior of the system (i.e., control step). For instance, developers in this step design object detection generators to perceive the environment and behavior generators to control the robot's motion. Next, we have the training step where developers configure the DRL agent, train it, and fine-tune its hyperparameters. Finally, developers repeatedly verify and update the system’s settings from the control and training steps to improve the agent's learning capabilities.

\textbf{DRL Deployment Lifecycle:} After verifying and testing the DRL agent, the system is ready for deployment by transferring the learned control policy to a physical or virtual platform for real-world use \cite{kurrek2020reinforcement}. A common approach is to deploy DRL systems on physical servers or on Cloud \cite{nair2015massively, liu2019lifelong, mao2019towards, noauthor_deploying_2020}. This approach offers developers tools and services like TensorFlow Serving \cite{olston2017tensorflow} or Amazon Sagemaker \cite{liberty2020elastic} to accelerate and facilitate deployment. 

In addition, other platforms for deploying DRL systems, such as mobile and embedded devices, are becoming popular~\cite{shiri2022e2hrl, jagannath2022mr}. However, practical-sized DRL agents cannot be deployed directly to these edge platforms because of their low computational power, memory size, and energy cost. To cope with limited edge platforms' resources, lightweight DL frameworks, such as TF Lite \cite{david2021tensorflow} and Core ML \cite{coreml}, are built to reduce the DNN footprint. These frameworks are also used in DRL to compress the agent’s DNN \cite{lall2023deep}. To decrease memory cost and processing overhead, TF Lite \cite{david2021tensorflow} and Core ML \cite{coreml}  employ model compression strategies such as quantization before deploying DRL models to edge platforms.

Furthermore, DRL systems may also be deployed in virtual platforms like browsers \cite{bamford2022griddlyjs, ma2019moving} and game engines~\cite{tucker2018inverse, kurrek2020reinforcement}. For browser deployment, developers employ particular frameworks and libraries for adapting DRL agents, such as TensorFlow.js and brain.js. Game engines, on the other hand, are frameworks conceived primarily for the design of video games. Popular game engines, such as Unity3d, provide neural network inference packages (e.g., Barracuda \cite{unitybarracuda}) that enable the usage of DRL agents within games. Consequently, developers may build DRL agents using frameworks like TensorFlow and PyTorch before deploying them in a game engine for inference.

This study analyses DRL deployment challenges in server/cloud, mobile/embedded system, browser, and game engine platforms, which host a large portion of DRL systems.

\section{Methodology}
\label{sec:methodology}
\begin{figure}[t]
  \centering
  \includegraphics[width=1.01\columnwidth]{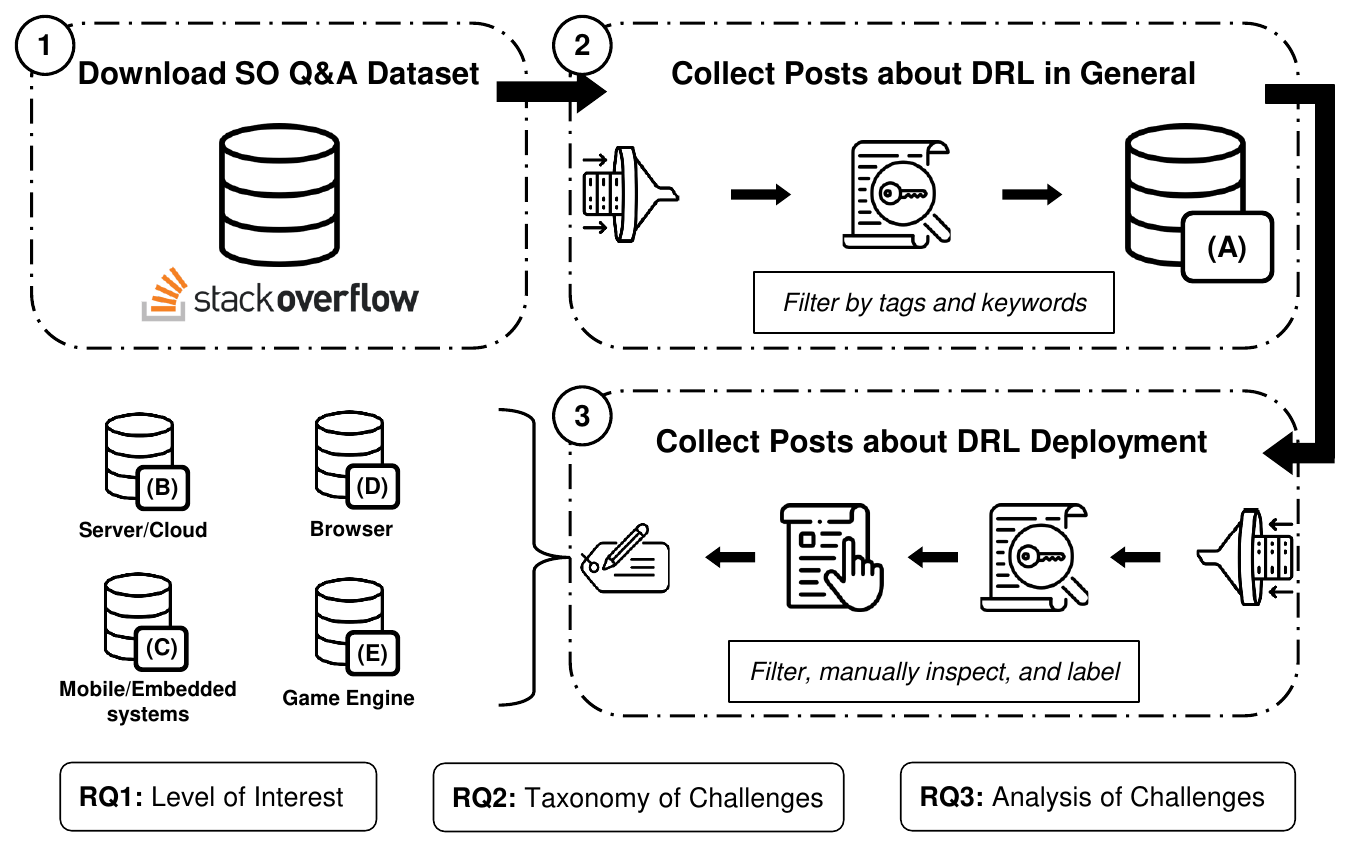}
  \caption{Overview of our study’s methodology.}
  \label{fig:wrokflow}
  \vspace{-1em}
\end{figure}

To better comprehend the challenges in deploying DRL-based systems, we analyze the relevant questions posted on SO.
Figure \ref{fig:wrokflow} highlights an overview of the three major steps of our methodology. We describe the steps we followed in our methodology in the rest of this section.

\noindent{\textbf{Step 1. Download the SO data dump.}} We download the SO dump from the official Stack Exchange Data Dump \cite{stackexchangearchive} on the 11th of October 2022. The dataset includes posts from July 2008 and each post contains information such as post type (i.e., question, answer, or wiki), creation date, tags, title, body, etc. Each question has one to five tags based on its topics and can have an accepted answer (meaning that the owner of the question has found a valid response to its question in one of the answers).

\noindent{\textbf{Step 2. Collect relevant DRL posts.}} To identify relevant posts, we manually search for tags/keywords based on our collective expertise (three researchers experienced in DRL) and qualitative search. During several discussion sessions, we collectively select representative tags and keywords for each subject. A similar methodology was used in previous studies \cite{study, cummaudo2020interpreting} effectively. Each time, we start with general tags/keywords (e.g., reinforcement learning) and add representative tags/keywords to enhance the tag-selection by manually searching for relevant, related terms based on the previously identified posts. In the following, we detail the steps to extract these posts. As a first step, we collect SO questions related to DRL in general. This dataset, denoted as \textit{A}, constitutes the starting point of our analysis. To build \textit{A}, we first extract questions tagged with “reinforcement-learning” or at least one of the top popular DRL frameworks, such as “stable-baselines” and “keras-rl” (the replication package has the full collection of tags \cite{replication}), and found 2996 relevant questions. Since many practitioners are still using classical DL frameworks such as Pytorch and TensorFlow to build DRL-based systems, we complement \textit{A} with questions tagged with “Pytorch” or “TensorFlow” and having a relevant keyword of DRL (the full collection of keywords can be found in the replication package \cite{replication}) within their title or body. Finally, we remove the duplicate questions and have retained a total of 3659 questions in \textit{A}.

\noindent{\textbf{Step 3. Collect relevant DRL deployment posts.}} Following similar studies \cite{study, guo2019empirical}, we select four representative deployment platforms of DRL systems to study, namely server/cloud, mobile/embedded system, browser, and game engine platforms. After filtering and manual analysis, we collected a dataset of 357 posts related to DRL deployment on all platforms. In the following, we detail the steps to extract these posts for each platform.

\noindent{\textbf{Server/Cloud Posts.}} We first define a vocabulary of words related to server/cloud platforms (such as, “cloud”, “server”, and “serving”). We then extract questions in \textit{A} having at least one word of Server/Cloud vocabulary in their title or body. We denoted this dataset as \textit{B}. To further complement \textit{B}, we extract SO questions having a DRL vocabulary word within their title or body and tagged with TF Serving, Google Cloud ML Engine, Azure ML, IBM Watson, or Amazon SageMaker. These tags represent popular cloud platforms for designing, training, and deploying ML systems and are used in similar studies \cite{study}. After manually removing duplicates and false positives (the manual analysis is detailed in the next paragraphs), we end up with 152 questions in \textit{B}.

\noindent{\textbf{Mobile/Embedded systems Posts.}} We first define a vocabulary of words related to mobile/embedded system platforms, like “mobile”, and “embedded system” (the replication package has the full collection of keywords \cite{replication}). We then extract questions in \textit{A} having at least one word of this vocabulary in their title or body. We denoted this dataset as \textit{C}. To further complement \textit{C}, we extract SO questions having a DRL vocabulary word within their title or body and tagged with at least one DL framework for mobiles/embedded systems such as TF Lite or one mobile/embedded hardware vendor like Arduino. After manually removing the duplicate and false positives, we end up with 69 questions in \textit{C}.

\noindent{\textbf{Browser Posts.}} We first define a vocabulary of words related to browser platforms (such as, “browser”). We then extract questions in \textit{A} having at least one word of browser vocabulary in their title or body. We denoted this dataset as \textit{D}. To further complement \textit{D}, we extract SO questions having a DRL vocabulary word within their title or body and tagged with tensorflow.js, brain.js, and ml5.js. These tags represent popular technologies used to deploy DL programs to browser platforms. After manually removing the duplicate and false positives, we end up with 45 questions in \textit{D}.

\noindent{\textbf{Game Engine Posts.}} We first define a vocabulary of words related to game engine platforms (such as, “game engine”). We then extract questions in \textit{A} having at least one word of game engine vocabulary in their title or body. We denoted this dataset as \textit{E}. To further complement \textit{E}, we extract SO questions having a DRL vocabulary word within their title or body and tagged with unity3d, ml-agent, barracuda, and game-engine. These tags represent popular game engine technologies. After manually removing the duplicate and false positives, we end up with 91 questions in \textit{E}.

\noindent{\textbf{RQ1: Level of interest.}} Following past study \cite{study}, we start by computing the number of questions linked to the DRL deployment per year to depict the evolution pattern of DRL deployment. The metrics are derived using datasets \textit{B}, \textit{C}, \textit{D}, and \textit{E} for each of the previous eight years (i.e., from 2015 to 2022). Second, to have an idea of the degree of difficulty of the DRL-based system deployment topic, we use two widely adopted metrics \cite{abdellatif2020challenges, alshangiti2019developing, ahmed2018concurrency}: the percentage of questions with no accepted answer \textit{(\%nAA)} and the response time needed to receive an accepted answer \textit{(RT)}. As a baseline for comparison, we used questions related to DRL but not related to deployment. To that aim, we remove the deployment-related questions (denoted as \textit{Dep}) (i.e., questions in \textit{B}, \textit{C}, \textit{D}, and \textit{E}) from the DL-related questions (i.e., questions in \textit{A}), and the remaining questions are referred to as non-deployment questions (denoted as \textit{Non-Dep}). In total, we had 2822 posts in \textit{Non-Dep}. Then, with a confidence level of 95\% and a confidence interval of 5\%, we randomly sampled posts from \textit{Non-Dep}. We randomly sampled since we needed to filter out false positive posts before starting the experiment and manual analysis on the full \textit{Non-Dep} was not practical. Our random sampling yields 339 posts in total. For the first measure (i.e., \textit{\%nAA}), we compute and compare the proportion of questions with no accepted response in \textit{B}, \textit{C}, \textit{D}, and \textit{E}, \textit{Dep}, and \textit{Non-Dep}. For the second measure (i.e., the time needed for an accepted answer), we choose the questions that have obtained accepted answers and then display the distribution and the median response time \textit{(MRT)} required to get an accepted answer for both deployment and non-deployment questions.

\noindent{\textbf{RQ2: Taxonomy of Challenges.}} We manually examine questions about DRL-system deployment to establish a taxonomy of challenges. The first two authors manually review all posts to eliminate duplicates and false positives. We include all posts related to DRL deployment and exclude false positive posts that address non-deployment concerns (like development). For taxonomy construction, we collected a dataset of 357 posts related to DRL deployment on all platforms. In the following, we present steps for the construction of the taxonomy.

\textit{Pilot construction and labeling.}  First, we randomly sample 40\% of the questions used for the taxonomy for a pilot construction of the taxonomy. The taxonomy for each kind of platform is constructed individually based on its corresponding samples. We follow an open coding procedure to inductively create the categories and subcategories of our taxonomy in a bottom-up way by analyzing all questions. The two first authors reviewed and revisited all of the questions to become acquainted with them. During this process, they carefully examine all aspects of each question, including the title, body, code samples, comments, responses, and tags. They then provide brief sentences as initial labels for the questions to illustrate the challenges underlying these questions. They then proceed to categorize the labels and develop a taxonomy of challenges in a hierarchical structure. This process is iterative as they travel back and forth between categories and questions to develop the taxonomy. All problems are discussed and resolved by introducing a third person, the arbitrator. The arbitrator has extensive expertise in DRL development and deployment, having published papers on DRL in top-tier journals and conferences. The arbitrator was also a senior research staff with 10+ years of experience as a researcher/practitioner in SE and RL. Finally, the arbitrator approves all of the taxonomy categories. After constructing the pilot, the two first authors independently label the remaining questions. Questions that cannot be categorized under the present taxonomy are placed in a new category called Pending, and the two first authors discuss whether new categories should be created. Cohen's Kappa metric for inter-rater agreement during independent labeling is 0.784, suggesting good agreement (posts labeled as “pending” were not included in the computation of Cohen’s Kappa). In situations where the two first authors could not agree, the post was handed over to the arbitrator to settle the labeling conflicts.

\textit{Taxonomy validation.} To guarantee that the final taxonomy is accurate and representative of realistic DRL deployment challenges, we validated it with a survey of DRL deployment practitioners/researchers. To attract recruiters, we used two alternative methods. First, we requested candidates via personal contacts. This resulted in a list of $11$ candidates who were contacted by email. We obtained $6$ positive responses from $4$ researchers and $2$ practitioners. Second, we leveraged GitHub and SO to gather information on potential survey respondents. To identify individuals with a strong grasp of DRL, we first identified the most popular RL frameworks on GitHub. We then retrieved and ordered contributors depending on their participation in the chosen repositories. To discover SO participants, we leveraged the posts retrieved during the mining phase. Following that, we identify the users who posted the question and answers to the selected posts. We searched the web for each identified person to locate their profile and emails from other sources, such as GitHub \cite{github} and Google Scholar \cite{googlescholar} since SO does not display user email addresses. This resulted in a second list of $207$ candidates who were contacted by email. We obtained $15$ positive responses from $9$ researchers and $6$ practitioners. Overall, we emailed the survey to $218$ individuals, and $21$ individuals responded ($13$ researchers and $8$ practitioners), resulting in a participation rate of $9.63\%$. The participant with the least experience had fewer than a year of practice in both ML/DL and DRL. The most experienced participant had more than 5 years of experience in both the ML/DL and DRL fields. The ML/DL and DRL experience medians were '3-5 years' and '1-3 years,' respectively. We utilized Google Forms \cite{googleforms}, a well-known online tool for data-collecting tasks, to build our survey form. We began the survey with background questions about job titles, DL/DRL experience, and familiar frameworks. Next, we moved on to the questions about our final taxonomy. We divided the survey into sub-categories and provided written descriptions for each category, including examples of its challenges. We provided the taxonomy (a simplified illustration), the name of each category, its description, and three questions related to each category. The first question was a "yes/no" question on whether the participant had ever experienced this challenge. If the answer is yes, we ask two further Likert-scale \cite{joshi2015likert} questions about the severity of the challenge and the needed effort to resolve it. Thus, we assessed not only the challenge's occurrence but also its severity as viewed by developers. In the final free-text question of our survey, we asked participants to name DRL deployment challenges they had and are not addressed in the taxonomy. This allowed us to determine if our taxonomy covered all developer challenges and what was missing. The survey questions are in the replication package \cite{replication}.

\noindent{\textbf{RQ3: Analysis of Challenges.}} After gathering the most common DRL deployment challenges that the SO community faces, the aim is to assess these challenges to identify the ones that are getting more traction and are harder for the DRL community to answer. First, we identify the most popular DRL deployment challenges among developers. To that end, we employ two measures of popularity that have been used in previous work \cite{abdellatif2020challenges, ahmed2018concurrency}: (1) The average number of views from both registered and unregistered users \textit{(AV)} of posts within a category, and (2) the average score \textit{(AS)} of posts within a category. \textit{AV} assesses community interest by showing how frequently posts are visualized in one category. The rationale behind this is that a post is popular among developers if many developers read it. \textit{AS} represents posts' recognized community value. Indeed, SO allows its users to up-vote posts that they find interesting and beneficial, and these votes are then combined to produce a score. After finding popular challenges, we assessed the difficulty of answering these challenges. Finding out if certain subjects are more difficult to answer than others will help us discover which challenges require greater community attention. It also helps us to indicate areas where improved tools/frameworks are needed to assist developers in tackling DRL deployment difficulties. To that end, we evaluate each challenge's difficulty using the two previously mentioned metrics: (1) \textit{(\%nAA)} and (2) \textit{(RT)}. Please see section \ref{sec:methodology}-RQ1 for further information on these two metrics.

\section{RQ1: Level of Interest}
\label{sec:RQ1}
\begin{figure}[t]
  \centering
  \includegraphics[width=0.8\columnwidth]{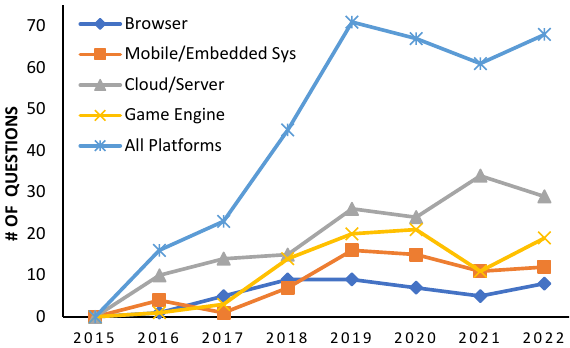}
  \caption{Number of Questions per Year.}
  \label{fig:fig1}
  \vspace{-1em}
\end{figure}

Figure \ref{fig:fig1} depicts the interest in deploying DRL systems measured by the number of questions on the SO. The graph shows that general interest in this subject is growing, confirming the study's relevance and importance.

Figure \ref{fig:fig1} illustrates a steady increase in posts discussing deploying DRL on servers/clouds. Moreover, the number of questions about mobile and embedded systems deployment grew considerably in 2018 compared to 2017. The reason is that some major vendors released their DL frameworks for mobile devices in 2017 (e.g., TFLite \cite{tensorflow2019tflite} and CoreML \cite{apple2017coreml}). We can also notice that the number of questions for deploying DRL systems on game engines rises steadily until 2020, when it begins to decline. 
Finally, we notice that number of questions about DRL deployment on browsers have not fluctuated since 2018. This can be explained by the release of TF.js \cite{tensorflowjs} in the same year. However, the number of browser deployment questions remains low compared to other platforms, indicating that DRL on browsers is still in its early stages. This low interest in deploying on browsers could be due to browsers' limited resources, whereas DRL agents are resource-intensive.

\begin{figure}[t]
  \centering
  \includegraphics[width=0.85\columnwidth]{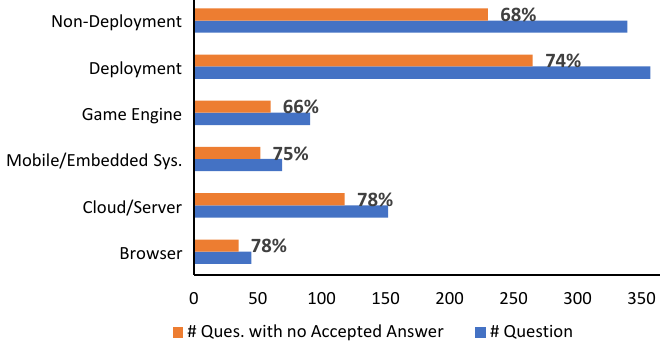}
  \caption{Number of Question with No Accepted Answer.}
  \label{fig:fig2}
\end{figure}

\begin{figure}[t]
  \centering
  \includegraphics[width=0.75\columnwidth]{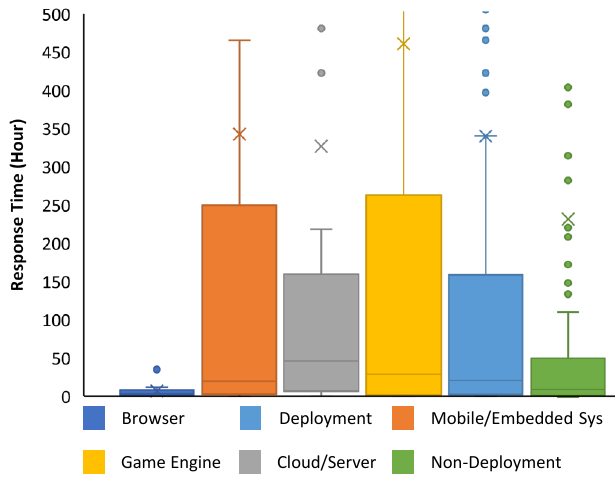}
  \caption{Time Needed To Receive an Accepted Answer.}
  \label{fig:fig3}
  \vspace{-1em}
\end{figure}

Figure \ref{fig:fig2} and \ref{fig:fig3} depict the difficulty of deploying DRL systems compared to other areas of DRL development. Figure \ref{fig:fig2} shows the ratio of questions with no accepted answer \textit{(\%nAA)} in DRL deployment and non-deployment-related questions. Whereas, Figure \ref{fig:fig3} shows the time needed to have an accepted answer \textit{(RT)} for DRL deployment- and non-deployment-related questions. Overall, both figures highlight that deployment-related questions are more difficult than non-deployment-related questions. Figure \ref{fig:fig2} shows that \textit{\%nAA} for DRL system deployment and non-deployment are 74\% and 68\%, respectively. This suggests that questions about DRL deployment are more difficult to answer than questions about other DRL issues. More specifically, \textit{\%nAA} for server/cloud, mobile, and browser platforms are 78\%, 75\%, and 78\% respectively. This suggests that deployment questions in these platforms are more difficult to answer compared to DRL non-deployment issues. However, deployment in game engines has a lower \textit{\%nAA} (66\%) than DRL non-deployment issues. 

Figure \ref{fig:fig3} shows the boxplot of the \textit{RT} required to get an accepted answer for DRL deployment and non-deployment-related questions. The median response time \textit{(MRT)} for deployment questions (22.58 hours) is 2.6 times that of non-deployment questions (8.83 hours), demonstrating that DRL deployment questions are more difficult to answer. Furthermore, the interquartile range (IQR) of \textit{RT} for non-deployment questions is 48.7 compared to 156 for deployment questions, indicating a higher spread for deployment questions. More specifically, IQR of \textit{RT} for server/cloud, mobile, and game engine platforms is 152.8, 247, and 261.3 respectively, as \textit{RT} in these platforms is more spread than \textit{RT} in DRL non-deployment questions. However, for DRL deployment on browsers, MRT and IQR are lower (3.1 and 7.4, respectively) than other platforms and non-deployment questions.

\begin{tcolorbox}[colback=gray!8,colframe=gray!40!black]
\textbf{Findings:} We found that questions about DRL deployment are increasing rapidly and gaining attention from the SO community. They are also more challenging to resolve than other issues of DRL system development.
\end{tcolorbox}

\section{RQ2: Taxonomy of Challenges}
\label{sec:RQ2}

\begin{table*}[htbp]
  \centering
  \caption{DRL deployment challenges and their categories. “Description” gives an overview of the category; “S/C”, "M/E", "G.E", "B" and "A" show the \% of each category in Server/Cloud, Mobile/Embedded, Game Engine, Browser, and All Platforms, respectively. 
  }
  \label{tab:taxonomy}
  \begin{tabular}{l l r r r r r}
    \hline
    \textbf{Challenge} & \textbf{Description} & \textbf{S/C}  & \textbf{M/E} & \textbf{G.E}  & \textbf{B} & \textbf{A}    \\
                       &                      & \textbf{(\%)} & \textbf{(\%)}  & \textbf{(\%)} & \textbf{(\%)} & \textbf{(\%)} \\
    \hline
    \textbf{General questions} &  &  \textbf{19.1}&  \textbf{18.8}&  \textbf{19.8}& \textbf{26.7} & \textbf{20.2} \\
    \hspace{4mm}\textit{Whole deployment process} &  Generic concerns regarding the entire deployment phase. &  5.3&  5.8&  5.5& 6.7& 5.6 \\
    \hspace{4mm}\textit{Conceptual questions} &  Concerns about DRL deployment concepts/background knowledge. &  9.9&  13&  12.1& 20& 12.3 \\
    \hspace{4mm}\textit{Limitations of platforms/frameworks} &  Limitations observed in DRL deployment platforms or frameworks. &  3.9&  -&  2.2& -& 2.2 \\
    \hline
    \textbf{Deployment infrastructure} &  &  \textbf{23}&  \textbf{17.4}&  \textbf{26.4}& \textbf{11.1} & \textbf{21.3} \\
    \hspace{4mm}\textit{Procedure} &  Concerns with process accomplishment (usually "how" questions).&  1.3&  10.1&  3.3& 4.4& 3.9 \\
    \hspace{4mm}\textit{Configuring the environment} &  Concerns related to preparing the deployment infrastructure.  &  7.2&  -&  -& -& 3.1 \\
    \hspace{4mm}\textit{Installing/setting libraries} &  Concerns about installing/building key libraries for the infrastructure. &  5.3&  7.2&  16.5& -& 7.8 \\
    \hspace{4mm}\textit{Managing resources} &  Concerns about managing/optimizing resources like RAM and CPU. &  2.6&  -&  -& -& 1.1 \\
    \hspace{4mm}\textit{Monitoring} &  Concerns about tracking important deployment infrastructure metrics. &  2&  -&  -& -& 0.9 \\
    \hspace{4mm}\textit{Multithreading} &  Concerns related to multithreading tasking in deployment. &  4.6&  -&  -& -& 2 \\
    \hspace{4mm}\textit{Library incompatibilities} &  Concerns about managing compatibility between libs while deploying. &  -&  -&  6.6& -& 1.7 \\
    \hspace{4mm}\textit{Asynchronous behavior} & Concerns about platforms' async. nature, e.g., Async/Await in browser. &  -&  -&  -& 6.7& 0.8 \\
    \hline
    \textbf{Data processing} &  &  \textbf{9.9}&  \textbf{13}&  \textbf{9.9}& \textbf{20}& \textbf{11.8} \\
    \hspace{4mm}\textit{Procedure} &  Concerns with data processing (usually "how" questions).&  2&  1.4&  2.2& 6.7& 2.5 \\
    \hspace{4mm}\textit{Setting size/shape of input data} &  Concerns about wrong size or shape of data during inference. &  7.9&  11.6&  7.7& 13.3& 9.3 \\
    \hline
    \textbf{RL environment} &  &  \textbf{13.8}&  \textbf{11.6}&  \textbf{11}& \textbf{24.4}& \textbf{14} \\
    \hspace{4mm}\textit{Procedure} &  Concerns with RL environment (usually "how" questions).&  0.7&  4.3&  5.5& 8.9& 3.6 \\
    \hspace{4mm}\textit{Configuring the RL environment} &  Concerns related to preparing and configuring the RL environment. &  5.3&  -&  3.3& -& 3.1 \\
    \hspace{4mm}\textit{Installing/setting libraries} &  Concerns about installing or building key libraries (e.g., Gym \cite{brockman2016openai}). &  7.9&  7.2&  -& 15.6& 6.7 \\
    \hspace{4mm}\textit{Library incompatibilities} &  Concerns about managing compatibility between libs while deploying. &  -&  -&  2.2& -& 0.6 \\
    \hline 
    \textbf{Communication} &  &  \textbf{4.6}&  \textbf{1.4}&  \textbf{12.1}& \textbf{4.4}& \textbf{5.9} \\
    \hspace{4mm}\textit{Procedure} &  Concerns with the communication process (usually "how" questions).  &  1.3&  -&  2.2& 2.2& 1.7 \\
    \hspace{4mm}\textit{Connection lost} &  Concerns with remote DRL components losing connection. &  3.3&  -&  5.5& -& 2.8 \\
    \hspace{4mm}\textit{Configuring remote settings} &  Concerns about settings up remote components (e.g., camera sensor). &  -&  -&  3.3& 2.2& 1.1 \\
    \hspace{4mm}\textit{Client/server interaction} &  Concerns about ensuring a remote client/server data interaction. &  -&  -&  1.1& -& 0.3 \\
    \hline
    \textbf{Agent loading/saving} &  &  \textbf{8.6}&  \textbf{14.5}&  \textbf{15.4}& \textbf{6.7}& \textbf{11.1} \\
    \hspace{4mm}\textit{Procedure} &  Concerns with the loading/saving process (usually "how" questions).&  4.6&  4.3&  6.6& -& 5.1 \\
    \hspace{4mm}\textit{Library incompatibilities} &  Concerns about managing compatibility between libs while deploying. &  3.9&  5.8&  3.3& -& 3.6 \\
    \hspace{4mm}\textit{Configuring frameworks/libraries} &  Concerns about fwk/lib configuration preventing agent loading/saving. &  -&  4.3&  3.3& -& 1.7 \\
    \hspace{4mm}\textit{Model conversion} &  Concerns about converting models from one format to another. &  -&  -&  2.2& -& 0.7 \\
    \hline
    \textbf{Performance} &  &  \textbf{6.6}&  \textbf{10.1}&  \textbf{5.5}& \textbf{-}& \textbf{6.2} \\
                         &  Category covering performance concerns while deploying DRL sys. &              &               &              &            \\
    \hline
    \textbf{Environment rendering} &  &  \textbf{13.2}&  \textbf{-}&  \textbf{-}& \textbf{-}& \textbf{5.6} \\
    \hspace{4mm}\textit{Procedure} &  Concerns with the rendering process (usually "how" questions).&  3.3&  -&  -& -& 1.4 \\
    \hspace{4mm}\textit{Configuring rendering components} &  Concerns regarding configuring libraries to allow rendering. &  5.9&  -&  -& -& 2.5 \\
    \hspace{4mm}\textit{Installing/setting libraries} &  Concerns about installing/building libraries for environment rendering. &  3.9&  -&  -& -& 1.7 \\
    \hline
    \textbf{Agent export} &  &  \textbf{-}&  \textbf{13}&  \textbf{-}& \textbf{-}& \textbf{2.5} \\
    \hspace{4mm}\textit{Procedure} &  Concerns with the agent exporting process (usually "how" questions).&  -&  4.3&  -& -& 0.8 \\
    \hspace{4mm}\textit{Model conversion} &  Concerns about converting models from one format to another. &  -&  5.8&  -& -& 1.1 \\
    \hspace{4mm}\textit{Model quantization} &  Concerns about quantization (e.g., quantized model in other fwks). &  -&  2.9&  -& -& 0.6 \\
    \hline
    \textbf{Request handling} &  &  \textbf{1.3}&  \textbf{-}&  \textbf{-}& \textbf{-}& \textbf{0.6} \\
                              &  Category covering concerns about making requests in the client side. &              &            &            &            \\
    \hline
    \textbf{Continuous learning} &  &  \textbf{-}&  \textbf{-}&  \textbf{-}& \textbf{6.7}& \textbf{0.8} \\
                                 &  Category covering CL’s concerns (learn continually without forgetting). & & & & \\
    \hline
  \end{tabular}%
  \vspace{-1em}
\end{table*}





Table \ref{tab:taxonomy} shows the taxonomy of DRL deployment challenges across four platforms. The taxonomy includes four sub-taxonomies that categorize challenges of deploying DRL systems to server/cloud, mobile/embedded devices, browsers, and game engine platforms. Each sub-taxonomy has two levels: the category of challenges (e.g., deployment infrastructure), and the challenge (e.g., monitoring). The taxonomy covers 11 unique categories and 31 unique challenges over the 4 platforms. Finally, our replication package includes the tree-structured taxonomy figures.

\subsection{Common Challenges to all platforms}

To prevent repetitions, we first present the common categories over all platforms and their challenges.

\subsubsection{General questions} This category includes general issues not limited to a particular deployment stage and includes three challenges.

\noindent{\underline{Whole deployment process.}} This challenge describes general concerns about the entire deployment phase. These are generally ``how'' questions, like ``How do I deploy the deep reinforcement learning neural network I coded in Pytorch to my website?" \cite{ref2}. Developers often ask about general guidelines to use in a specific case (e.g., \cite{ref3}). Answers usually provide tutorials, documentation-like material, or a list of generic steps to achieve the desired goal. This challenge has 5 to 7\% of questions on all platforms. 

\noindent{\underline{Conceptual questions.}} This challenge includes questions on fundamental concepts or background knowledge regarding DRL deployment, such as ``What is the easiest 2D game library to use with Scala?" \cite{ref4}. It accounts for 9.9\%, 13\%, 20\%, and 12.1\% of questions in server/cloud, mobile/embedded systems, browser, and game engines respectively. This suggests that even the fundamentals of DRL deployment can be difficult for developers.

\noindent{\underline{Limitations of platforms/frameworks.}} This challenge is about the Limitations of deployment platforms and frameworks. For example, in this post \cite{ref5}, the SO user reported a Unity ml-agents problem. The engineer assigned to this bug apologizes for the inconsistent documentation, which was later rectified.

\subsubsection{Data processing} This category discusses the difficulties encountered while shaping raw data into the format required by the deployed DRL system. This category accounts for 20\%, 13\%, 9.9\%, and 9.9\% of the browser, mobile/embedded system, game engine, and cloud/server questions respectively. In the following, we list common challenges in this category.

\noindent{\underline{Procedure.}} Procedure covers questions about a specific deployment task, as opposed to the "Whole deployment process" which covers questions about the whole deployment phase. "Unity ML Agent, CameraSensor - Compression Type" \cite{ref6} is an example in which the SO user asked about the model input when using the unity-trained model in a real-world setting. In the remainder of the paper, we do not duplicate the Procedure descriptions in other categories due to the page limits.

\noindent{\underline{Setting size/shape of input data.}} Setting the size/shape of data is a typical challenge in data pre-processing. When the input data has an unexpected size/shape during the inference step, improper behavior occurs. In this SO post \cite{ref7}, for example, the user is attempting to take a model built by Unity ML-Agents and run inferences using Tensorflow, where she encountered a shape mismatch error.

\subsubsection{Deployment infrastructure} This category includes concerns in preparing the deployment infrastructure for DRL systems. This category has the largest ratio of questions in game engine platforms and cloud/server platforms, with 26.4\% and 23\%, respectively. It also accounts for 17.4\%  and 11.1\% of questions in mobile/embedded systems and browsers respectively. The challenges shared by all platforms in this category are: 

\noindent{\underline{Configuring the environment.}} When deploying DRL systems, developers must set up several environment variables, paths, and configuration files, all of which have numerous choices, making configuration difficult. Problems that arise during this stage are addressed under this challenge.

\noindent{\underline{Installing/setting libraries.}} Furthermore, developers must install or set essential libraries to prepare the deployment infrastructure. This type of concern is addressed in this challenge.

\noindent{\underline{Library incompatibilities.}} Some developers have trouble using libraries while deploying DRL systems. Indeed, the continuous growth of libraries makes version compatibility of frameworks/libraries difficult for developers. For example, a SO post reported an error that was triggered by Tensorflow 1.7.0 having an incompatible version with the Unity3d ml-agents used version \cite{ref8}.

\subsubsection{RL environment} This category includes challenges in preparing the RL environment for deployment. It comprises nearly the same challenges as the Deployment infrastructure category. However, we consider a question related to the RL environment when a wrong behavior occurs during attempting to prepare the RL environment, rather than the deployment infrastructure. The Deployment Infrastructure topic covers broader issues related to the deployment ecosystem. For instance, an issue with configuring Docker containers for deployment is considered a Deployment Infrastructure challenge \cite{chakraborti_pyglet_2018}, whereas a concern with configuring the Gym environment is considered an RL environment challenge \cite{dhanush-ai1990_cannot_2018}. We find a noticeable variation in patterns between these two categories after establishing this distinction. With a question ratio of 24.4\%, the RL environment category is the second-highest among browser platforms. It also accounted for 11.6\%, 11\%, and 13.8\% of questions in the mobile/embedded system, game engine, and cloud/server platforms, respectively.

\subsubsection{Communication} This category addresses issues with communication between DRL system components. For example, in this SO question titled ``How to ensure proper communication between a game in C++ and an RL algorithm in Python?" the user asks about a method to ensure proper communication between a TensorFlow/Keras-based RL agent and a C++ game. This category accounts for 4.4\%, 1.4\%, 12.1\%, and 4.6\% of the browser, mobile/embedded system, game engine, and cloud/server questions respectively. 

\subsubsection{Agent loading/saving} This category contains challenges with saving a DRL agent from one platform/framework and reloading it in another platform/framework. First, developers may encounter difficulties converting models from one format to another in order to use them on a specific platform (e.g., \cite{ref10}). Furthermore, incorrect setup of a certain framework or library may hinder the process of loading or storing a DRL agent to the required platform. In AWS Sagemaker, for example, a faulty setup prohibited a user from recovering an agent's trained DNN \cite{ref11}. Finally, incompatibility across frameworks or libraries may pose an extra barrier when loading or storing a DRL agent \cite{ref12}. The category of agent loading/saving accounts for 6.7\%,14.5\%, 15.4\%, and 8.6\% of questions in the browser, mobile/embedded system, game engine, and cloud/server platforms respectively.

\subsubsection{Performance} When deploying DRL systems, performance is a critical factor that developers must address. Execution time, latency, and hardware consumption are all critical factors that could affect the DRL system in production. This category includes all performance issues that may arise while deploying DRL systems. These issues might arise primarily in two places: the environment and the DRL agent. In one SO post \cite{ref14}, for example, the user is attempting to run a large number of simulated environments in parallel using Google Cloud Kubernetes Engine. However, the simulation on her development device is twice faster than the one on Kubernetes. In another SO post \cite{ref15}, the user is attempting to benchmark two Google Coral \cite{coral} devices for the deployment of a DRL agent. She was disappointed since the Coral device is significantly slower than the CPU. This category comprises 10.1\%, 5.5\%, and 6.6\% of questions in mobile/embedded systems, game engines, and cloud/server platforms, respectively.

\subsection{Other Challenges in Browser}

\noindent{\textit{Continuous Learning.}} Continuous Learning (CL) is the model's ability to continually learn and adjust its behavior in real time. DRL systems' trial-and-error nature makes CL the go-to approach for adjusting the DRL agent to rapidly changing conditions in production. This category addresses concerns regarding using CL in production with DRL systems. In this SO post, \cite{ref16}, for example, the user asks about a method to continually adapt a trained bot to emulate a real player. This category accounts for 6.7\% of browser platform questions.

\subsection{Other Challenges in Mobile/Embedded System}

\noindent{\textit{Agent Export.}} Agent export can be achieved by (1) directly converting its trained model into the required format or by (2) using dedicated frameworks to transform the model to a format that runs on the deployment platform. In mobile/embedded system platforms, agent export accounts for 13\% of all questions. Agent exporting is essential when deploying DRL agents on edge platforms (e.g., mobiles) due to their limited resources and different operating systems. In the following, we discuss challenges within this category.

\noindent{\underline{Model Conversion.}} It includes issues associated with any setting misbehavior or incorrect use of model conversion frameworks (e.g., ONNX) \cite{ref17}.

\noindent{\underline{Model Quantization.}} Quantization lowers the accuracy of model parameters’ representation in order to minimize the memory footprint of DNNs. In this challenge, developers often struggle with (1) combining quantization frameworks like TF Lite with other frameworks or platforms \cite{ref18} or (2) working with varied precision floating points \cite{ref19}.

\subsection{Other Challenges in Server/Cloud}

\subsubsection{Request handling} This category includes issues with making requests on the client side and accounts for 1.3\% of cloud/server platform questions. Developers struggle mainly with customizing the request body \cite{ref13} and obtaining information on serving models through request.

\subsubsection{Environment rendering} This category discusses issues with rendering the environment while running DRL systems on a server or in the cloud. It accounts for 13.2\% of cloud/server questions (third highest). Developers, in this category, struggle to configure frameworks and/or libraries to allow rendering in headless servers \cite{ref20, ref21}.

\subsection{Survey Results}

Table \ref{tab:validation_survey} presents the validation survey findings. We show the percentage of participants who replied ``yes" to whether they faced related challenges. We also show the severity of each challenge category and the reported effort needed to fix it. The survey participants have faced all sorts of challenges, proving the taxonomy's relevance. With 81\% approval, ``RL environment" is the most approved category. 65\% and 41\% of these participants said this category has a high severity and requires high effort, respectively. 19\% of survey participants had encountered ``Continuous Learning" and ``Agent Export", the least-approved categories. Yet, 50\% of these participants believe these categories have a high severity and require high effort. The average approval rate for common challenges to all platforms is 61.2\%, indicating that the final taxonomy matches the experience of most participants. The remaining platform-specific challenges average 28.5\% "yes" responses. This may be because participants only work on one platform and don't have experience with all platforms. Finally, some participants provided challenges they thought were not in the taxonomy: \textit{Train/production RL environment gap:} One participant highlighted the challenge of matching the training and deployment RL environment. Indeed, this problem is frequent in DRL deployment literature (e.g., the sim-to-real gap \cite{zhao2020sim}). 
\textit{Partial observability \& non-stationarity:} Another participant commented that non-stationarity makes DRL deployment difficult \cite{dulac2019challenges}. For example, recommendation systems in production must adapt to changing user behavior. 
Finally, one participant's comment was remarkable. 
According to her/him, one of the biggest deployment issues is aligning and expressing DRL-specific challenges (e.g., sim-to-real gap \cite{zhao2020sim}) with teams who don't have much RL/ML knowledge. DRL is different from common engineering practices, making team communication difficult.


\begin{tcolorbox}[colback=gray!8,colframe=gray!40!black]
\textbf{Findings:} We identified 31 challenges, most of which are common across deployment platforms. The most common challenges are related to the deployment infrastructure, while the RL environment challenge is confirmed by most survey participants. 
\end{tcolorbox}

\begin{table}[t]
  \centering
  \caption{Validation Survey Results}
  \label{tab:validation_survey}
  \resizebox{0.48\textwidth}{!}{%
  \begin{tabular}{l|c |c c c|c c c}
    \hline
    \textbf{Challenge} & \textbf{Resp.} & \multicolumn{3}{c|}{\textbf{Severity (\%)}} & \multicolumn{3}{c}{\textbf{Effort Required (\%)}} \\
     & Yes(\%) & Low & Med. & High & Low & Med. & High\\ 
    \hline
    RL Env. & 81 & 12 & 24 & 65 & 12 & 47 & 41 \\ 
    Deployment Infra. & 48 & 20 & 60 & 20 & 40 & 30 & 30 \\ 
    Data Processing & 76 & 44 & 38 & 19 & 38 & 44 & 19 \\ 
    Communication & 24 & 20 & 60 & 20 & 20 & 80 & 0 \\ 
    Agent Sav./Load. & 71 & 40 & 40 & 20 & 53 & 27 & 20 \\ 
    Performance & 67 & 14 & 43 & 43 & 21 & 43 & 36 \\ 
    Continuous Learn. & 19 & 0 & 50 & 50 & 0 & 50 & 50 \\ 
    Agent Export & 19 & 25 & 25 & 50 & 25 & 25 & 50 \\ 
    Env. Rendering & 52 & 27 & 36 & 36 & 18 & 64 & 18 \\ 
    Req. Handling & 24 & 0 & 100 & 0 & 20 & 60 & 20 \\
    \hline
  \end{tabular}%
  }
  \vspace{-1em}
\end{table}


\section{RQ3: Analysis of Challenges}
\label{sec:RQ3}

\begin{table*}[htbp]
  \centering
  \caption{The popularity/difficulty per topic.}
  \label{tab:pop_diff}
  \resizebox{\textwidth}{!}{%
  \begin{tabular}{l|c c c c|c c c c|c c c c|c c c c|c c c c}
    \hline
    \textbf{Challenge} & \multicolumn{4}{c|}{\textbf{Browser}} & \multicolumn{4}{c|}{\textbf{Mobile/Emb. Sys}} & \multicolumn{4}{c|}{\textbf{Game Engine}} & \multicolumn{4}{c|}{\textbf{Cloud/Server}} & \multicolumn{4}{c}{\textbf{All Platforms}} \\
    
    & \multicolumn{2}{c}{\textbf{POP.}} & \multicolumn{2}{c|}{\textbf{DIFF.}} & \multicolumn{2}{c}{\textbf{POP.}} & \multicolumn{2}{c|}{\textbf{DIFF.}} & \multicolumn{2}{c}{\textbf{POP.}} & \multicolumn{2}{c|}{\textbf{DIFF.}} & \multicolumn{2}{c}{\textbf{POP.}} & \multicolumn{2}{c|}{\textbf{DIFF.}} & \multicolumn{2}{c}{\textbf{POP.}} & \multicolumn{2}{c}{\textbf{DIFF.}}\\
     & AV & AS & \%nA & MRT & AV & AS & \%nA & MRT & AV & AS & \%nA & MRT & AV & AS & \%nA & MRT & AV & AS & \%nA & MRT\\ 
    \hline
    \rowcolor{gray!20}
    General Question & 486& 0.6& 67& 2.4& 740& 0.8& 85& 2.5& \textbf{2713}& \textbf{9.1}& 78& 25.5& 354& 0.7& 79& 8.1 & 1035 & \textbf{2.8} & 78 & 3 \\
    \rowcolor{gray!20}
    Data Processing & 203& 0.7& 78& 7& 215& 0.7& 68& 64.9& 376& 0.7& 78& 1.5& 1362& 1.1& 53& 21.6 & 657 & 0.8 & 67 & 14.2 \\
    \rowcolor{gray!20}
    Deployment Infra. & 711& \textbf{1}& 80& 0.8& \textbf{1986}& 2& 92& 0.1& 1159& 1.1& 63& 14.4& 791& 0.7& 83& 45.6 & 1091 & 1.1 & 78 & 37.6 \\
    \rowcolor{gray!20}
    RL Env. & 579& 0.6& 91& 0.2& 1198& -0.1& 75& 10.3& 533& 1.4& 50& 326.9& 2089& 2.4& 81& 32.9 & \textbf{1303} & 1.4 & 76 & 32.9 \\
    \rowcolor{gray!20}
    Communication & \textbf{1215}& 0.5& \textbf{100}& N.A& 86& 0& \textbf{100}& N.A& 452& 0.4& \textbf{82}& 14.5& 2330& 2.6& 71& 192.9 & 1133 & 1.1 & \textbf{81} & \textbf{98} \\
    \rowcolor{gray!20}
    Agt Sav./Loading & 189& 0.3& 67& \textbf{34.8}& 817& 0.9& 50& \textbf{340}& 1289& 1.1& 50& 258.9& 725& 1.8& 69& 71.9 & 905 & 1.2 & 58 & 76 \\
    \rowcolor{gray!20}
    Performance & -& -& -& -& 952& \textbf{3.4}& 71& 84& 1391& 2& 60& \textbf{474.1}& 693& 1.2& 70& 50 & 975 & 2 & 68 & 50 \\
    \rowcolor{gray!50}
    Cont. Learning & 235& 0.7& 67& 6.8& -& -& -& -& -& -& -& -& -& -& -& - & 235 & 0.7 & 67 & 6.8 \\
    \rowcolor{gray!50}
    Agt Export & -& -& -& -& 775& 0.9& 78& 10.4& -& -& -& -& -& -& -& - & 775 & 0.9 & 78 & 10.4 \\
    \rowcolor{gray!50}
    Env. Rendering & -& -& -& -& -& -& -& -& -& -& -& -& \textbf{8039}& \textbf{7.8}& 90& \textbf{244.4} & 8039 & 7.8 & 90 & 244.4 \\
    \rowcolor{gray!50}
    Req. Handling & -& -& -& -& -& -& -& -& -& -& -& -& 69& 0& \textbf{100}& N.A & 69 & 0 & 100 & N.A \\

    \hline
    \textbf{All Challenges} & 473 & 0.6 & 78 & 3.1 & 982 & 1.1 & 75 & 19.7 & 1268 & 2.6 & 66 & 28.8 & 1946 & 2.1 & 78 & 45.8 & 1401 & 1.9 & 74 & 20.6 \\

    \hline
  \end{tabular}%
  }
\end{table*}












Table \ref{tab:pop_diff} shows the popularity and difficulty of RQ2's categories of challenges. Popularity is expressed by the Average Views \textit{(AV)} and Average Scores \textit{(AS)} metrics, while difficulty is expressed by the percentage of questions with No Accepted Answer \textit{(\%nAA)} and the median time (in hours) to receive an accepted answer \textit{(MRT)}. Challenges in Table \ref{tab:pop_diff} are categorized by the deployment platforms. Platform-common and platform-specific challenges are highlighted in light and dark gray, respectively. 

\noindent{\textbf{Common Challenges to all platforms.} Across all platforms, the most difficult challenge in terms of \textit{\%nAA} and \textit{MRT} is “Communication”. “Communication” has also the second highest \textit{AV}, making it a popular and difficult challenge in DRL deployment. The posts on this challenge, however, have the second-worst \textit{AS} in all platform-common challenges. This high \textit{\%nAA} combined with low \textit{AS} may indicate that these posts' questions are unclear or poorly written which decreases the likelihood of an accepted answer. 

The most popular challenge in terms of \textit{AV} is the “RL environment”. This challenge has also the third-highest \textit{AS}, which highlights the importance and popularity of this challenge. Furthermore, in our validation survey, ``RL environment" was the most approved challenge. Survey participants agreed that RL environment challenges in production (e.g., sim to real gap \cite{zhao2020sim}, non-stationarity \cite{dulac2019challenges}) are prevalent.

In addition, “General Question” is the most popular challenge in terms of \textit{AS}. Interestingly, this category has the lowest \textit{MRT}. This suggests that even general, easy-to-answer, questions about DRL deployment are often requested.

On the other hand, “Data Processing” has the lowest \textit{AV} and the lowest \textit{AS}. Also, it has the second-lowest \textit{MRT} and the second-lowest \textit{\%nAA}, making this challenge the least popular and second-least difficult. To understand the reason behind this behavior, we examine the posts on this challenge and find that most posts are asking about simple tensor shaping problems. Also, Data Processing is closely related to traditional DL challenges, which could explain why the SO respondents tend to answer this challenge faster. 

\noindent{\textbf{Challenges specific to one platform.} “Environment Rendering”, a specific challenge to Cloud/Server platforms, is the most popular challenge with the highest \textit{AV} and the highest \textit{AS}. This challenge is also the one with the highest \textit{MRT} and the second-largest \textit{\%nAA}. This demonstrates that the “Environment Rendering” challenge is among the most popular and difficult challenges in cloud/server-related challenges. The other platform-specific challenges, like “Continuous Learning” and “Request Handling”, are less popular and less difficult compared to the other challenges. This can be explained by the fact that these challenges are very specific and are not faced by developers on a regular basis.

\noindent{\textbf{Correlation analysis.} Finally, to have a full view of the DRL deployment challenges, we examined if there is a statistically significant correlation between the difficulty and popularity of the assessed challenges. In particular, we use the Spearman Rank Correlation Coefficient \cite{zar1972significance} to verify the correlations between the two popularity metrics (\textit{AV}, and \textit{AS}) and the two difficulty metrics (\textit{\%nAA} and \textit{MRT}). We choose Spearman’s rank correlation since it does not have any assumption on the normality of the data distribution. Results show a statistically significant correlation between \textit{AS} popularity metric and the \textit{MRT} difficulty metric (coeff=0.47, p-value=0.011). This demonstrates that highly scored questions need more time to receive an accepted answer, and difficult challenges tend to be popular among developers.
Our replication package \cite{replication} has further details about correlation analysis. 



\begin{tcolorbox}[colback=gray!8,colframe=gray!40!black]
\textbf{Findings:} Across all platforms, RL environment-related challenges are the most popular whereas communication-related challenges are the most difficult. 
Overall, we observe a significant positive correlation between the popularity and difficulty level of the challenges. 
\end{tcolorbox}

\section{Implications}
\label{sec:Implications}
\begin{figure}[t]
  \centering
  \includegraphics[width=1.0\columnwidth]{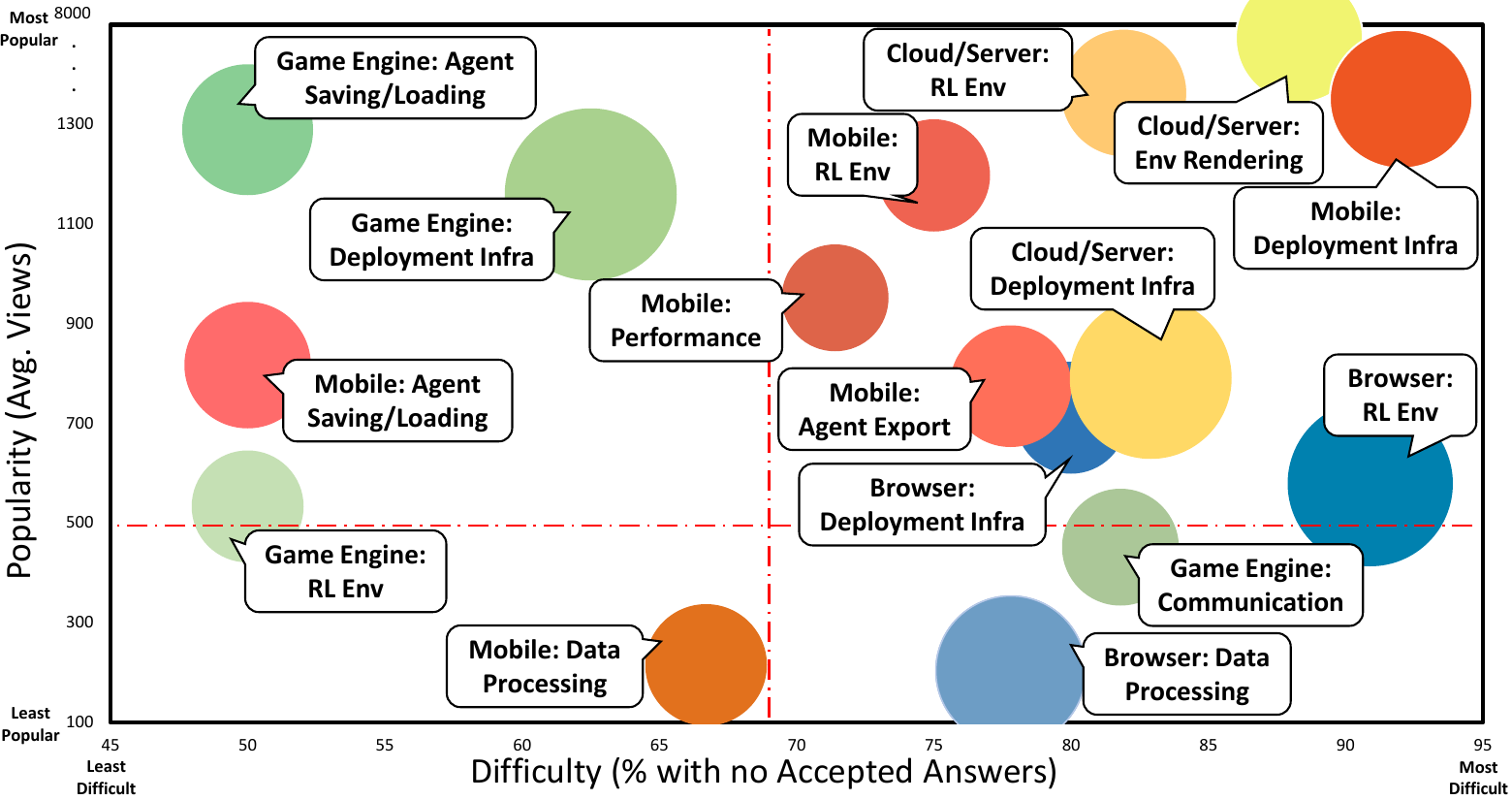}
  \caption{Bubble plot of topic popularity and difficulty.}
  \label{fig:fig6}
  \vspace{-1em}
\end{figure}


We describe, in the following, how our findings may provide insights to practitioners, researchers, and framework developers to improve the deployment of DRL systems. Figure \ref{fig:fig6} illustrates a bubble plot ranking the challenges in terms of their popularity and difficulty. The bubble's size indicates the proportion of posts for a challenge. The figure's four quadrants show the challenge's level of popularity and difficulty. $AV$ serves as a proxy for popularity, while \%$nAA$ serves as a proxy for difficulty. Finally, due to the large number of challenges (31), we removed challenges with less than 10\% of posts on each platform.

\noindent{\textbf{Implication for Practitioners.}} Despite being the most popular challenge, questions about ``RL environment" remain mostly unresolved (see Figure \ref{fig:fig6}). In fact, unlike vanilla DL systems, deploying DRL systems requires a focus on several components like the environment, the agent, and the communication between them. This finding should be embraced by the community to improve tutorials and documentation in order to reduce the burden of deploying DRL systems. Our findings can also assist developers in better prioritizing their effort by addressing the most challenging topics in DRL deployment. Software managers can similarly account for this by allocating more resources and time to DRL-specific tasks.  
Finally, DRL deployment is based on the interaction between DRL and Software Engineering. As a result, DRL deployment necessitates developers that are knowledgeable in both domains, making it a difficult task. Our taxonomy may be used as a checklist for developers, encouraging them to obtain the essential skills before deploying DRL systems.

\noindent{\textbf{Implication for Researchers.}} As revealed in our study, DRL deployment is growing in popularity among developers, but developers face a wide range of challenges, many of which are DRL-specific (e.g., RL environment challenges). Moreover, as seen in Figure \ref{fig:fig6}, DRL-specific challenges are particularly hard to solve. Thus, researchers are urged to propose approaches and solutions to assist developers in addressing these deployment challenges, such as Configuration. Several challenges in our taxonomy are connected to configuration since DRL incorporates many interacting components thus many configurable elements. This insight can inspire researchers to provide automated configuration solutions to make various deployment tasks easier for developers. In addition, rule-based verification may be introduced, as they are quite useful for detecting and diagnosing misconfigurations. Likewise, strategies for monitoring and notifying developers about potential issues throughout the deployment process might be introduced. Monitoring the deployment process is a difficult problem in DRL \cite{feit2022explaining, garcia2015comprehensive, dulac2019challenges} that requires paying attention to the variety of potential root causes, which include hardware and software failures, as well as concept and data drifts \cite{padakandla2020reinforcement}. 

\noindent{\textbf{Implication for Framework Suppliers.}} According to our findings, many developers struggle with the whole deployment process. Indeed, developers frequently highlight the poor quality or the lack of documentation in these questions (e.g., SO post \cite{ref5}), indicating that the documentation should be enhanced. Additionally, the predominance of the ``Conceptual questions" category (which hit 20\% of posts in Browser) implies that framework suppliers should increase their documentation's completeness, especially given that DRL deployment necessitates a diverse range of knowledge and expertise. Figure \ref{fig:fig6} also demonstrates that the ``Deployment infrastructure" is the most prevalent challenge in our taxonomy in terms of number of posts. This figure also shows this challenge's difficulty and popularity, as it appears in the figure's most popular/difficult quadrant on three of four platforms. These findings may be used to encourage better and more intuitive end-to-end DRL frameworks. Currently, depending on the deployment platform, incremental deployment, automating/optimizing data processing pipelines, and using serving systems for models are popular coping strategies that developers use. However, end-to-end frameworks such as MLflow \cite{mlflow} are gaining popularity in the ML community and we are starting to witness initiatives aiming at ``productionizing" DRL utilizing these technologies (e.g., Spark \cite{rlbakery}, MLflow \cite{confcasttv}). Nevertheless, the majority of these frameworks do not pair well with DRL or do not support it at all. We believe that our study might push toward better and improved DRL-specific end-to-end frameworks.


\section{Threats to Validity}
\label{sec:thre-valid}

One potential threat is the selection of specific tags and keywords to identify relevant posts. Our automatic collection process may be biased by the pre-selected tags and keywords. To mitigate this threat, we chose popular frameworks and platforms to ensure representativeness. However, the keyword-matching collection process may include false positives or exclude posts without pre-selected keywords.

Another potential threat is the reliance on SO as a single data source for studying developer challenges. While we retrieved a fair number of relevant posts, it is possible that we overlooked valuable insights from other sources. However, we believe that our results are still relevant owing to the diversity of SO users, including both novices and experts.  

Finally, the subjectivity of the manual analysis is a potential threat to the validity of our work. To mitigate this risk, the first two authors individually inspect posts and reached an agreement. They also had the assistance of a third expert author when there is a conflict. The inter-rater agreement is substantial confirming the labeling procedure's reliability.


\section{Related Work} 

\label{sec:related-work}
In recent years, ML has solved various real-world problems. Yet, the deployment of ML models remains challenging. Numerous software engineering (SE) studies have addressed these challenges to help practitioners. Breck et al. \cite{breck2017ml} provided an actionable checklist of 28 tests and monitoring criteria to assess ML system production readiness. Similarly, Paleyes et al. propose practical consideration by evaluating reports on deploying ML systems (including RL). 

Recently, researchers focused on DL-specific deployment challenges across several platforms. Cummaudo et al. \cite{cummaudo2020interpreting} examined developer challenges with computer vision services (i.e., APIs). Ma et al. \cite{ma2019moving} assessed seven JavaScript-based DL frameworks on Chrome and found that DL in browsers is still in its early stage. Xu et al. \cite{xu2019first} proposed the first empirical study on DL in Android applications. Guo et al. \cite{guo2019empirical} examined the performance gap of DL models when moved to mobile devices and Web browsers, whereas Openja et al. \cite{openja2022empirical} evaluated ONNX and CoreML for deploying DL models. Finally, Chen et al. \cite{study} examined the challenges of DL deployment across three platforms. 

Despite all these efforts, DRL-specific deployment challenges remain unstudied. Instead, few SE research like \cite{nikanjam2022faults} examined DRL development challenges. Unlike these works, our study focused on DRL deployment to bridge the knowledge gap between research and practice. The current DRL deployment research is focused on addressing the problem from an academic perspective. Panzer and Bender \cite{panzer2022deep} presented a systematic literature review of DRL deployment in production and provide a global overview of DRL applications in various production system domains. In addition, Dulac-Arnold et al. \cite{dulac2019challenges} list nine specific challenges (e.g., partial observability, and non-stationarity) to productionize RL to real-world situations. For each challenge, they propose literature-based methodologies and metrics for evaluation. These studies complement ours since they use literature to extract and understand DRL deployment challenges in production, while we focus on practitioners and the industry.

\section{Conclusion}
\label{sec:conclusion}
This work proposes an empirical study on SO to understand the challenges practitioners faced when deploying DRL-based systems. We examined $357$ SO posts about DRL deployment and found that the general interest in DRL deployment is growing. Our findings also reveal that DRL deployment is harder than other parts of DRL systems development, pushing us to investigate its specific challenges. To that end, we build a taxonomy of $31$ unique challenges for deploying DRL to server/cloud, mobile/embedded systems, game engines, and browsers. DRL deployment has unique challenges like data processing (managing data pipelines), RL environment, loading/saving agent, and continuous learning. This uniqueness stems from DRL’s design and learning paradigm. Unlike DL and conventional software, deploying DRL systems requires focus on the environment, agent, and communication between them. We hope this study stimulates future research and helps the community solve DRL-based system deployment's most prevalent and difficult challenges. In future work, we plan to broaden our data sources and interview experts and practitioners to further confirm and expand our findings.


\balance

\section*{Acknowledgment}
\label{sec:ack}
This work is funded by the Fonds de Recherche du Quebec (FRQ), the Canadian Institute for Advanced Research (CIFAR), and the National Science and Engineering Research Council of Canada (NSERC). However, the findings and opinions expressed in this paper are those of the authors and do not necessarily represent or reflect those organizations/companies.

\bibliographystyle{IEEEtran}
{\footnotesize
\bibliography{refs.bib}}
\end{document}